\documentclass[letterpaper, 10 pt, conference]{ieeeconf}  
\IEEEoverridecommandlockouts                             
\usepackage{cite}
\usepackage{amsmath,amssymb,amsfonts,amsthm}
\DeclareMathOperator*{\argmin}{argmin} 

\newtheorem{definition}{Definition}
\newtheorem{proposition}{Proposition}
\newtheorem*{remark}{Remark}
\usepackage{algorithmic}
\usepackage{graphicx}
\usepackage{textcomp}
\usepackage{subcaption}
\usepackage{comment}
\usepackage{xcolor}
\def\BibTeX{{\rm B\kern-.05em{\sc i\kern-.025em b}\kern-.08em
    T\kern-.1667em\lower.7ex\hbox{E}\kern-.125emX}}

\usepackage{epstopdf}

\begin{document}
\title{The Quasi-static Kinematic Modeling of Tracked Vehicles for Urban Terrain\\}

\author{Anushri Dixit$^{1}$, Joel Burdick$^{1}$
\thanks{*This work was funded in part by DARPA, under the Subterranean Challenge program.}
\thanks{$^{1}$Anushri Dixit and Joel Burdick are with Control and Dynamical Systems at California Institute of Technology, Pasadena, CA 91125, USA
        {\tt\small adixit@caltech.edu}, {\tt\small jwb@robotics.caltech.edu}}%
}
\maketitle

\begin{abstract}
This paper develops a new quasi-static modeling framework for tracked robots. Given a set of track speeds, this our model predicts the vehicle's instantaneous rigid body motion. We introduce three specific models: a model for tracked operation on flat ground, a model for vehicle motion when the track's grouser tips touch hard ground, and a model for operation on stairs. Experiments show that these models predict tracked vehicle motion more accurately than existing kinematic models, and predict phenomena which are not captured by other models. These novel models provide a basis for new feedback control approaches.
\end{abstract}


\section{Introduction} \label{sec:intro}

Most mobile robots are propelled by wheels.  However, tracked robots can have better mobility over uneven terrain, and they can potentially climb complex structures, such as stairs and rubble, that are impassable for wheeled robots \cite{brunner_motion_2012}.  While the complex interaction between the track and its supporting terrain provides superior traction, tracked propulsion is more difficult to model, and potentially to control.  A key issue is that, except for forward motion, portions of the track must be sliding over the terrain as the vehicle moves.  This paper revisits the modeling of tracked vehicle motion using new methods.  These models can lead to newer control approaches and better performance.  We also provide a new model and insights for tracked operation on stairs.

There is a long history of research on tracked vehicle modeling \cite{bekker_off_1962,bekker_introduction_1969,wong_general_2001}.  Early work endeavored to understand the mechanisms of traction \cite{kitano_analysis_1977}, or the physical factors that affect the ability of a tracked vehicle to turn  \cite{jozaki_steerability_1979}, \cite{baladi_analysis_1981}.  Naturally, the analysis and modeling of the skidding process has received considerable attention, as a greater understanding can lead to better autonomous control of tracked vehicle motion \cite{wang_design_1990, gonzalez_localization_2009,jingang_yi_kinematic_2009,wei_yu_analysis_2010,janarthanan_lateral_2011,gonzalez_autonomous_2014,pentzer_model-based_2014}.  The slipping process has also be considered from the viewpoint of tracked vehicle power use and efficiency \cite{tianyou_guo_simplified_2013}. The majority of these previous models can be characterized as {\em terramechanics models}. This paper develops a control-oriented mechanics modeling framework.


  There are many practical approaches to tracked vehicle motion planning and control. Shiller was one of the first to study a dynamic model of tracked vehicles \cite{ZviShiller} for purposes of motion planning. Shiller  conceptualized the skid steer process as a dynamic nonholonomic constraint, which can then be used as a constraint in the motion planning process.  While it represents a real advancement, Shiller's work also relies upon the prediction of track forces given vehicle motion.  Others have used on-line estimation to estimate some of the kinematic or dynamic model parameters \cite{le_estimation_nodate, yi_adaptive_2007, moosavian_experimental_2008, dar_slip_2010}.

The authors of \cite{martinez_approximating_2005} derived a kinematic equivalence between skid-steer and wheeled differential-drive models. They computed the instantaneous centers of rotation (ICR) of the individual tracks at low speeds. This idea is the closest to our work. We use the Power Dissipation Methodology which is a kinematic reduction of the Lagrangian model (and equivalently the Newtonian model) to compute the kinematics of the vehicle. In \cite{morales_power_modeling_2009}, the authors further modeled power losses due to dynamic friction and used this power model to minimize the energy spent for navigation tasks on flat ground. Our work provides a generalized framework for kinematic analysis of tracked robots given the robot geometry and the contact model at any instant (not limited to planar motion). We are able to obtain instantaneous body velocity from track speeds {\em without requiring experimental identification}. 

Since tracks are one of the few practical alternatives to legs for stair climbing, several works have analyzed their operation and agility on stairs.  These efforts have typically focused on the forces and geometry involved in stair climbing \cite{yugang_liu_track--stair_2009}. Others have managed the real-time control of stair climbing using simple differential drive equations  \cite{yalin_xiong_vision-guided_2000,mourikis_autonomous_2007, Steplight},  Though simple enough to implement, we will show that these equations do not accurately portray vehicle motion on stairs.   This paper gives one of the first methods to derive feedback equations of tracked vehicles on stairs.


This paper contains the following contributions.  First,  we apply  the {\em Power Dissipation Methodology} (PDM) to the modeling of tracked vehicles to yield a quasi-static model that estimates vehicle motion given the track speeds as input. Such models can support advances in feedback control design for tracked vehicles.  We focus on a quasi-static model because we are most interested in cases where a tracked vehicle negotiates complex environments, which typically occurs at low speeds.  

Second, when a tracked vehicle operates on flat homogeneous ground, our proposed model with Coulomb friction predicts that the vehicle will move {\em exactly} like a differential drive vehicle, but with a fictitious wheel radius that depends {\em nonlinearly} on the track length and width.  This result justifies the simplified kinematic modeling approaches that have previously been used, but it improves them by giving a rigorous model (that is based on track length and width) for the lumped parameters coefficients that have previously been derived in ad-hoc ways. 

Third, using the PDM, we model tracked vehicle motion on stairs.  Importantly, we show that a model which predicts vehicle motion on flat ground gives erroneous predictions on stairs.  Moreover, our model shows that as the vehicle climbs stairs, particularly with a nonzero angular speed, the changes in supporting contacts result in a switching behavior as the tracks gain and lose contact with stairs. This phenomena is replicated in our experiments.

We also apply the stair modeling framework to the study of tracks equipped  with {\em grousers}.   Grousers improve traction in soil \cite{li_research_2014}.  However, when a tracked vehicle travels over hard ground, only the grouser tips contact the terrain surface.  Our model captures this situation with high fidelity.

Finally, we experimentally verify our theoretical predictions with data gathered from a {\em Rover Robotics Flipper} tracked vehicle while it drives on flat ground and while it climbs stairs. The models derived predict the vehicle's motion more accurately than other simplified modeling methods and confirm the stair climbing motion predictions.

The next section reviews the PDM modeling method.  Section \ref{sec:model} applies this method to a tracked vehicle, implicitly yielding a novel set of input-output equations. Section \ref{ssec:grousers} analyzes the case of grousers on hard flat ground, while Section \ref{ssec:stairs} specializes the model to vehicle operation on stairs.  Section \ref{sec:experiments}  presents experimental results on flat ground as well as on stairs, and compares the results to our model predictions. 

\section{Background on the Power Dissipation Method} \label{sec:background}

A {\em quasi-static} system is one in which the inertial forces are negligible, that is, either $m$ or $a$ is negligible in $F = ma$. Such an approximation of a system is useful when the dissipative forces are much greater than the inertial forces. Quasi-static models are often useful in practice, as they can abstract key relationships in a useful form.

\subsection{Minimum Power Principle}\label{ssec:PDM}

The principles of quasi-static modeling in this paper go back many decades.  However,  within the field of robotics, Peshkin and Sanderson\cite{minPower} proposed the following \textit{principle of minimum power} for quasi-static systems:

\textit{A quasi-static system chooses that motion, from among all motions satisfying the constraints, which minimizes the instantaneous
power. }

The above principle does not hold for all quasi-static systems. It holds for forces that are parallel to the velocities of the system's particles, but that  are independent of those velocity magnitudes (for example, Coloumb frictional forces). It also holds for forces that are independent of particle velocity (e.g., gravitational forces). Hence, the principle can be applied to the quasi-static sliding of a system of particles. 

Alexander and Maddocks\cite{WMR} introduced a similar notion for rolling motions in kinematic models of wheeled mobile robots and showed that these motions represent the least power dissipation through frictional forces. This method is termed the {\em Power Dissipation Methodology} (PDM).  Murphey and Burdick \cite{PDM} considered the PDM from a control theoretic perspective.  They determined the conditions under which a quasi-static model derived from the PDM is a rigorous {\em kinematic reduction} of a full Lagrangian model. 

\subsection{Power Dissipation Methodology}
 We assume that the system configuration, $q = (q_g, q_r) \in Q$, consists of states $q_g$ and control inputs $q_r$. $Q$ is written as a product of the state and control manifolds $Q_g$ and $Q_r$ respectively.   The Power Dissipation Method is based on the notion of the {\em Power Dissipation Function}.  

\vskip 0.05 true in
\begin{definition} \label{defn:PDF}
Given a system with configuration $q = (q_g, q_r)\in Q_g \times Q_r = Q$ and $q_r$ fixed, the {\bf power dissipation function}, $\mathcal{D}((q_g,q_r)) ((\dot{q}_g, \dot{q}_r))$ models the amount of power dissipated due to the motions, $\dot{q}_g$, of the system's particles at configuration $q_g$ while the inputs $\dot{q}_r$ are fixed.
\end{definition}
\vskip 0.05 true in

For tracked vehicles, the power dissipation occurs when portions of the track slide over the terrain. The PDM is based on the minimization of the power dissipation function:

\vskip 0.05 true in
\begin{proposition}
Given a power dissipation function, $\mathcal{D}$, as defined in Definition. \ref{defn:PDF}, the system’s motion, $\dot{q}_g$ at any given instant is the one that minimizes $\mathcal{D}$ with respect to $\dot{q}_g$ while the inputs $q_r$ are held fixed:
\begin{equation}\label{eq:minD}
    \dot{q}^*_g = \argmin_{\dot{q} \in T_{q_g}Q_{q_g}} \mathcal{D}((q_g, q_r))((\dot{q}_g, \dot{q}_r))
\end{equation}
\end{proposition}

\vskip 0.05 true in
A rigorous analysis of this methodology can be found in \cite{PDM}.
Practically, the PDM yields an input-output relationship $\dot{q}_g = h(\dot{q}_r)$.  Note that the function $h(\cdot)$ can be discontinuous, and set-valued at some system configurations.

Consider a rigid body system that contacts its surroundings at multiple points. The dissipation function quantifies the power lost while the system overcomes frictional contact forces during its motion. This paper considers Coulomb frictional forces, but other models are possible.  If a contact does not slip, its relative velocity is 0. If it slips, the relative velocity is given by $\omega(q)\dot{q}$ for some function $\omega(q)$. If $N_i$ denotes the normal force at the $i^{th}$ contact  and $\mu_i$ is the friction coefficient, the dissipation function for $\kappa$ contacts is:
\begin{equation}
    \mathcal{D}(q)(\dot{q}) =\sum_{i = 1}^{\kappa} \mu_iN_i|\omega(q)\dot{q}|
\end{equation}

We use this principle to obtain kinematic models of tracked robot motion on flat homogeneous ground and then on stairs. In the next section, we outline the derivations of the dissipation function for different scenarios. In Section \ref{sec:analysis}, we then derive models for each of these scenarios by numerically solving \eqref{eq:minD}.

\section{Model} \label{sec:model}
\begin{figure}[htbp]
\vskip -0.1 true in
\centering
    \begin{subfigure}[t]{0.24\textwidth}{
    \centerline{\includegraphics[width=35mm,scale=0.4]{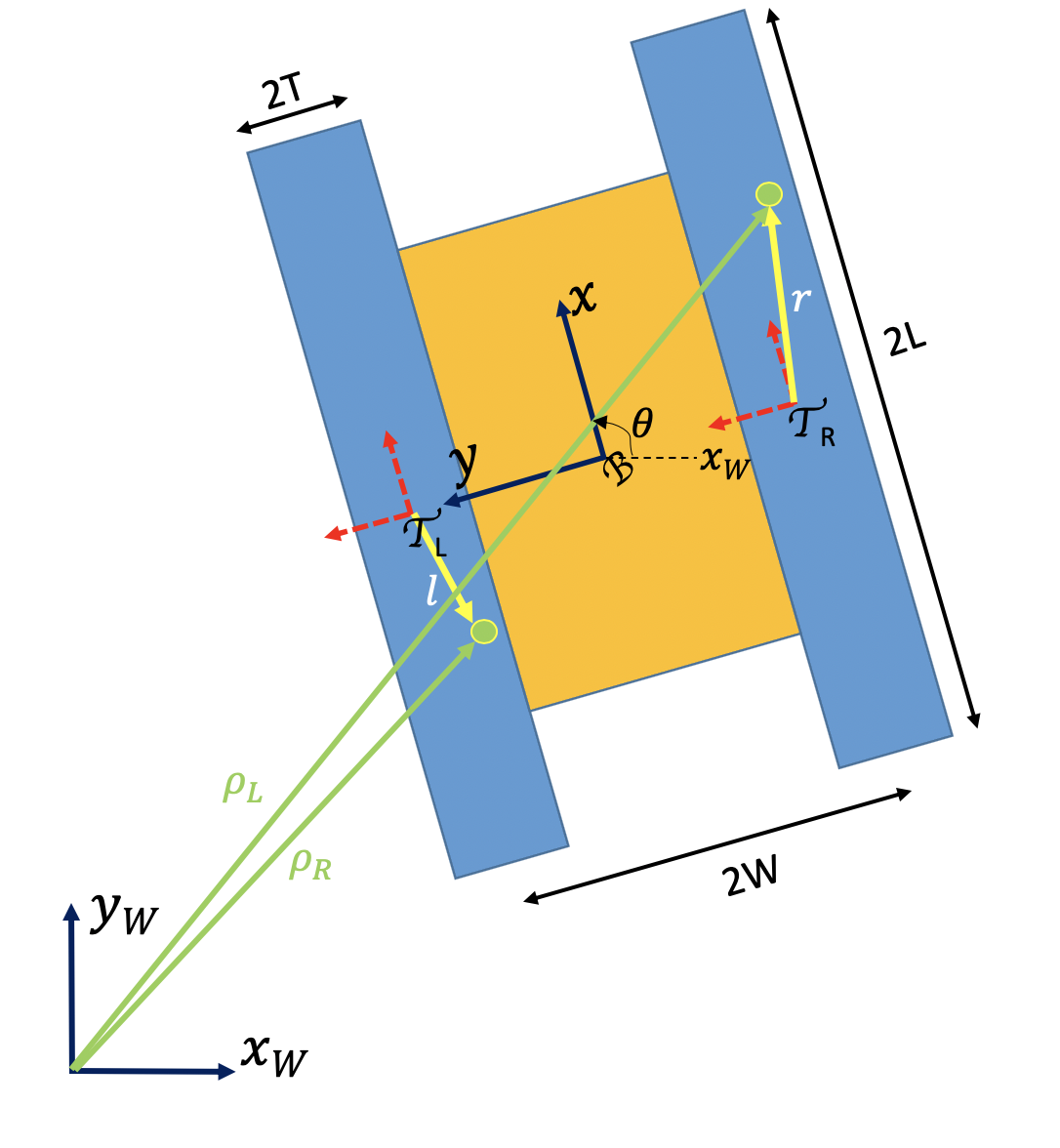}}
    \vskip -0.05 true in
    \caption{Top view}
    \label{fig:model}}
    \end{subfigure}%
    \begin{subfigure}[t]{0.24\textwidth}{
    \centerline{\includegraphics[width=40mm,scale=0.4]{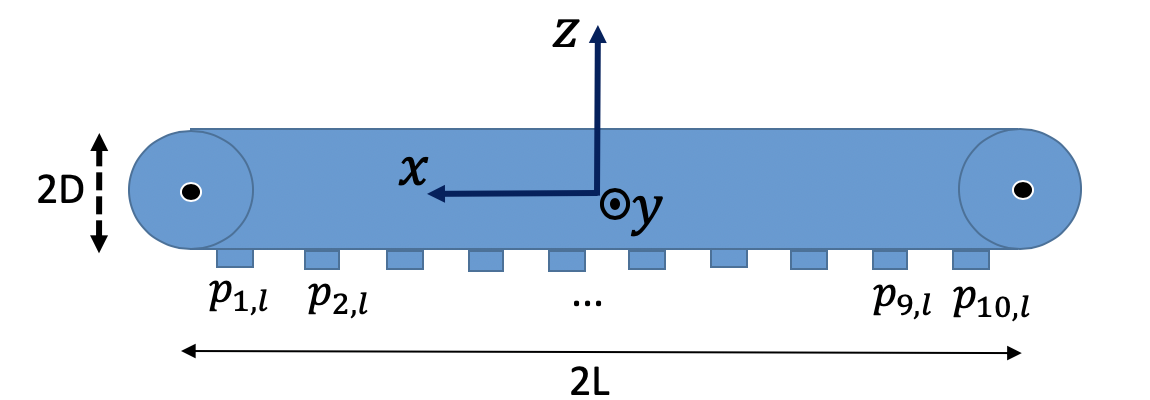}}
    \vskip -0.05 true in
     \caption{Side view}
    \label{fig:grousers_fig}}
    \end{subfigure}
    \caption{Vehicle Geometry: top and side view (with grousers)}
\end{figure}

Consider a tracked vehicle model shown in Fig. \ref{fig:model}. We assume a {\em symmetric} vehicle with identical tracks.  A right-handed body fixed reference frame, $\mathcal{B}$, is fixed to the center of symmetry.  Each track has length $2L$ and width $2T$.  The tracks are assumed to be driven by a sprocket with diameter $2D$.  The distance between the track center lines is $2W$. The origin of $\mathcal{B}$, is located at a height $D$ above the ground plane.  Its $x$-axis points in the forward driving direction, the $y$-axis points to the left, and the $z$-axis points out of the plane.

We seek to compute the power dissipation function, Definition \ref{defn:PDF}, for the vehicle in Fig. \ref{fig:model}.  To describe local coordinates on each track surface, define references frames $\mathcal{T}_L$ and $\mathcal{T}_R$ (Fig. \ref{fig:model}) that are oriented parallel to $\mathcal{B}$, with their origins located on the ground plane, a distance $\pm W$ from the origin of $\mathcal{B}$ along $\mathcal{B}$'s $y$-axis.  In these local reference frames, each point on the right and left track surface (that are in contact with the ground) respectively has coordinates $\begin{bmatrix} r_x & r_y\end{bmatrix}$ and $\begin{bmatrix} l_x & l_y\end{bmatrix}$.  
In the body-fixed frame, points on the track surfaces are located at:
\begin{equation*}{\small
    \rho_{r}^B = \begin{bmatrix} r_x \\ -W + r_y \\ -D\end{bmatrix} \ \ \ \ \ \ 
    \rho_{l}^B = \begin{bmatrix}l_x \\ W + l_y \\ -D\end{bmatrix} }
\end{equation*}
When the vehicle is on flat ground moving with velocity $\begin{bmatrix} \dot{x} & \dot{y} & \dot{\theta}\end{bmatrix}^T$ the velocities of the points on the left and right treads are (in frame $\mathcal{B}$):
\begin{equation} \label{eq:vr}{\small
    v_{r} = v_{r}^B = \begin{bmatrix} \dot{x} \\ \dot{y}\end{bmatrix} + \dot{\theta}\begin{bmatrix}
    W - r_y \\ r_x\end{bmatrix} - S_r\begin{bmatrix} 1 \\ 0\end{bmatrix}}
\end{equation}
\begin{equation}\label{eq:vl}{\small
    v_{l} = v_{l}^B = \begin{bmatrix} \dot{x} \\ \dot{y}\end{bmatrix} + \dot{\theta}\begin{bmatrix}
    -(W + l_y) \\ l_x\end{bmatrix} - S_l\begin{bmatrix} 1 \\ 0\end{bmatrix}}
\end{equation}
The right and left track speeds are denoted $S_r$ and $S_l$.  Their signs are positive when the tracks move toward the rear of the vehicle (in a manner to propel the vehicle forward):
\begin{equation*}
    S_r = -\dot{r}_x, \quad S_l = -\dot{l}_x, \quad \dot{r}_y = \dot{l}_y = 0.
\end{equation*}
The total dissipated power is the integral over the track surfaces of the power dissipated at each point where the track touches the ground.  According to the power dissipation methodology, the vehicle’s rigid body velocity due to given track inputs $S_r$ and $S_l$ is the one which which minimizes the total dissipated power when $S_r$ and $S_l$ are held fixed:
\begin{equation}\label{eq:1}
\begin{aligned}
    \begin{bmatrix} \dot{x} \\ \dot{y} \\ \dot{\theta}\end{bmatrix} =  \argmin_{\dot{x}, \dot{y}, \dot{\theta}} &  \int \limits_{\rho_r \in \Omega_r } \alpha_{r} \|v_r\|\mathrm d\rho_r +  \int \limits_{\rho_l \in \Omega_l} \alpha_{l} \|v_l\|\mathrm d\rho_l
\end{aligned}
\end{equation}
where the tracks contact the ground in regions $\Omega_r$ and $\Omega_l$ and $\alpha_{r} = \mu(\rho_r) N(\rho_r) $ and $\alpha_{l} = \mu(\rho_r) N(\rho_r) $.

\subsection{Modeling on flat ground} \label{ssec:flat_ground}

\begin{figure}[htbp]
\vskip -0.1 true in
\centering
    \begin{subfigure}[t]{0.24\textwidth}{
        \centering
        \includegraphics[width=40mm,scale=0.4]{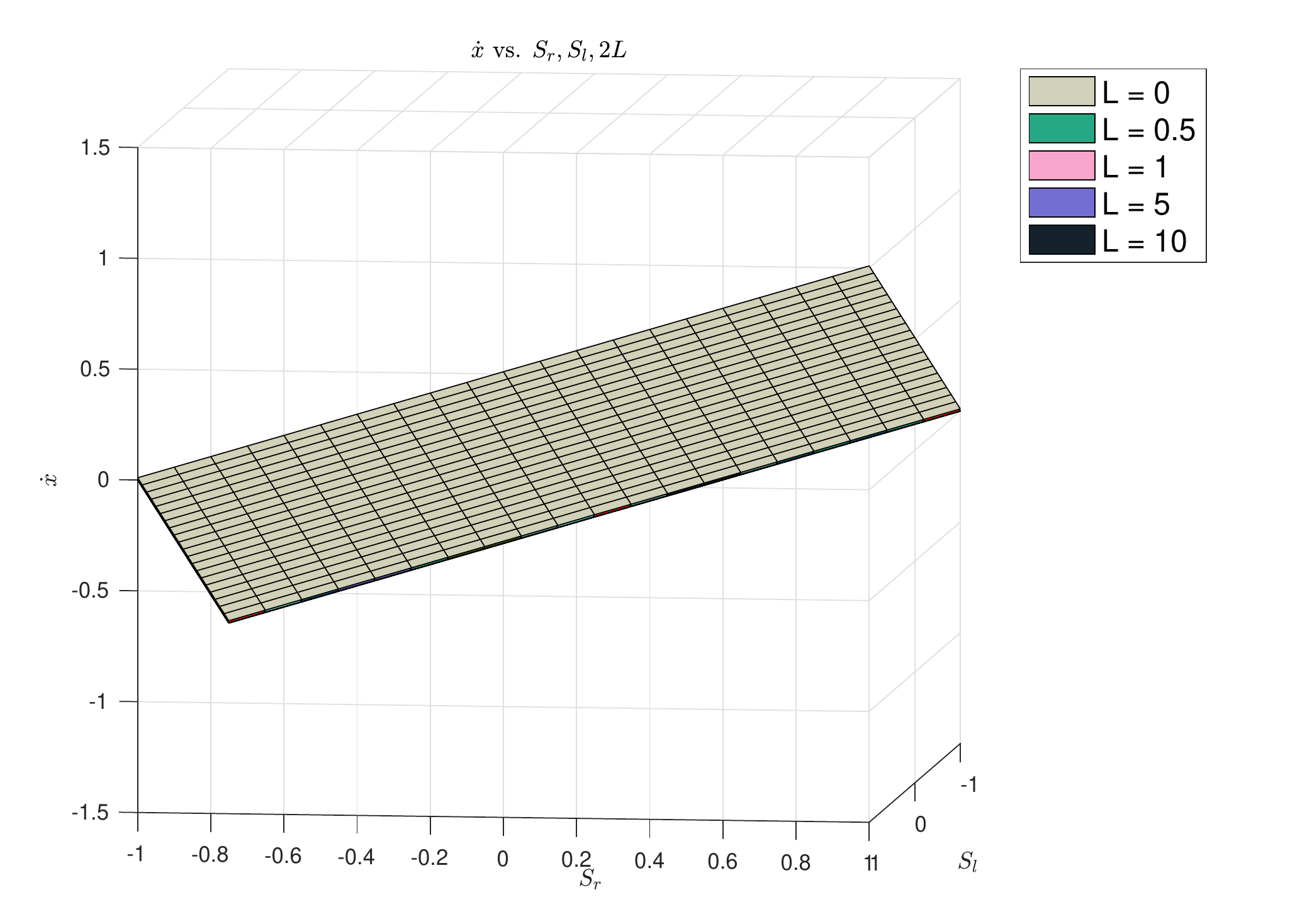}
        \caption{$\dot{x}$ vs. $S_r, S_l$ for different L}
        \label{fig:xdot}}
    \end{subfigure}%
    \begin{subfigure}[t]{0.24\textwidth}{
        \centering
        \includegraphics[width=40mm,scale=0.4]{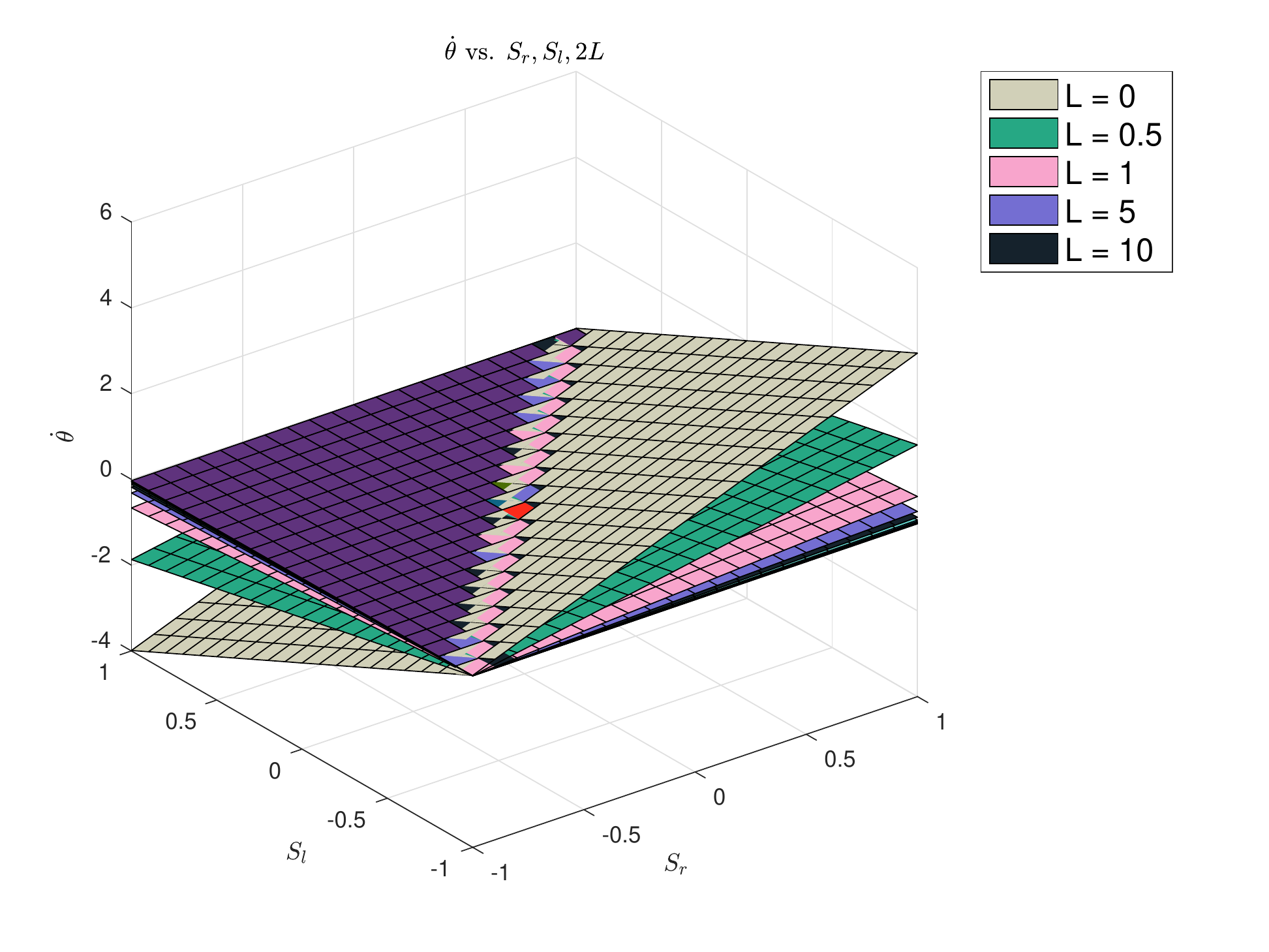}
        \caption{$\dot{\theta}$ vs. $S_r, S_l$ for different L.}
        \label{fig:tdot}}
    \end{subfigure}
\caption{Flat ground simulation: Variation in linear and angular velocity}
\label{fig:fig2}
\end{figure}

 It is impractical to solve (\ref{eq:1}) in general.  Consider a simpler case where the vehicle drives on flat, solid, homogeneous ground.  Let us reasonably assume that vehicle weight is evenly distributed across the two tracks, and that ground pressure is uniformly distributed across the treads: $ N(\rho_r) = N(\rho_l) = \text{constant} \quad \forall \, \rho_r \in \Omega_r, \rho_l \in \Omega_l$.  We also assume a Coulomb friction model for track/ground contact, with a uniform friction coefficient across tracks: $\mu_r = \mu_l = \text{constant} \quad \forall \, \rho_r \in \Omega_r, \rho_l \in \Omega_l$. 
Substituting these simplifications and (\ref{eq:vr}), (\ref{eq:vl}) into the integrand of (\ref{eq:1}) yields the power dissipation function (in a body-fixed frame):

\begin{equation*}
{\small
\begin{aligned}
&\mathcal{D}(\dot{x}, \dot{y}, \dot{\theta}, S_r, S_l) = \int \limits_{\rho_r \in \Omega_r } \|v_r\|\mathrm d\rho_r  + \int \limits_{\rho_l \in \Omega_l} \|v_l\|\mathrm d\rho_l \\
= & \int \limits_{-T}^{T}\int \limits_{-L}^{L} \sqrt{\left(\dot{x} + \left(W-r_y\right)\dot{\theta}-S_r\right)^2 + \left(\dot{y} + r_x\dot{\theta}\right)^2}\,\mathrm dr_x \mathrm dr_y \\
& +\int \limits_{-T}^{T} \int \limits_{-L}^{L} \sqrt{\left(\dot{x} - \left(W+l_y\right)\dot{\theta}-S_l\right)^2 + \left(\dot{y} + l_x\dot{\theta}\right)^2}\,\mathrm dl_x \mathrm dl_y 
\end{aligned}}
\end{equation*}
\normalsize
The quasi-static equations of motion are obtained by minimizing $\mathcal{D}$ with respect to $\begin{bmatrix} \dot{x} & \dot{y} & \dot{\theta}\end{bmatrix}^T$ for a given $S_r$ and $S_l$. This double integral can be computed numerically on a modest computer in faster-than real time.   It can be similarly minimized with respect to body velocities in real-time using common minimization procedures. 
\begin{remark}
Fig.~\ref{fig:fig2} shows the values of $\dot{x}, \, \dot{\theta}$ for varying lengths of track and track speeds. Note that $\dot{y}$ is excluded in Fig.~\ref{fig:fig2} because $\dot{y}=0$ regardless of vehicle dimensions and track speeds when operating on flat ground. Note that the relationship between $\dot{x}$ and $S_r, S_l$ does not vary as $L$ varies. The value of $\dot{\theta}$, however, varies as $\mathrm{C}(S_r - S_l)$, C is constant depending upon track geometry. When $L = 0$, C~$=$~$\frac{1}{2W}$, i.e, the vehicle behaves like a differential drive robot. A similar analysis can be done for the track width ($2T$). To the authors' knowledge, no such dependence of angular velocity on track length and width has been derived theoretically in the literature previously. 
\end{remark}

\subsection{Modeling of grouser effects on flat ground} \label{ssec:grousers}
Most tracked vehicles employ grousers, Fig. \ref{fig:grousers_fig}. On flat, hard surfaces, only the tips of the grousers make ground contact. Hence, the power dissipation function becomes an integral over a set of line contacts:
\begin{equation*}
{\small
\begin{aligned}
&\mathcal{D}_{g}(\dot{x}, \dot{y}, \dot{\theta}, S_r, S_l) \\
=&\sum \limits_{n=1}^{N_c}\int \limits_{-T}^{T} \alpha_{n,r} \sqrt{\left(\dot{x} + \left(W - r_y\right)\dot{\theta}-S_r\right)^2 + \left(\dot{y} + p_{n,r}\dot{\theta}\right)^2} \mathrm{d}r_y\\
+& \, \sum \limits_{n=1}^{N_c} \int \limits_{-T}^{T}\alpha_{n,l}\sqrt{\left(\dot{x} - \left(W+l_y\right)\dot{\theta}-S_l\right)^2 + \left(\dot{y} + p_{n,l}\dot{\theta}\right)^2}\mathrm{d}l_y
\end{aligned}}
\end{equation*}
where $\alpha_{n,r} = \mu N(p_{n,r})$ and $\alpha_{n,l} = \mu N(p_{n,l})$. Note that $p_{n,r/l}$ denotes the $x$-component (in the body frame) of the $n^{th}$ grouser contact between the right/left track and the ground.  Similarly, $N(p_{n,r/l})$ denotes the normal force at $p_{n,r/l}$, which is assumed to be uniform across the width of the grouser.  One can arrive at this formula by substituting into the double integral a set of delta functions located at the distances where the grousers contact the ground. 

The normal forces are calculated from a moment balance that balances the vehicle's weight. The normal forces vary, based on where the grousers contact the ground. We assume the friction coefficients to be the same on all grouser tips. 
 Note that the grouser positions on the right and left tracks may not be synchronized, so that the body velocity varies with changes in the relative grouser displacements. However, in practical vehicle designs, this variation is not significant. Section \ref{sec:analysis} describes an approximate closed form solution to this case.

\subsection{Modeling on stairs} \label{ssec:stairs}

\begin{figure}[htbp]
\centerline{\includegraphics[width=35mm,scale=0.5]{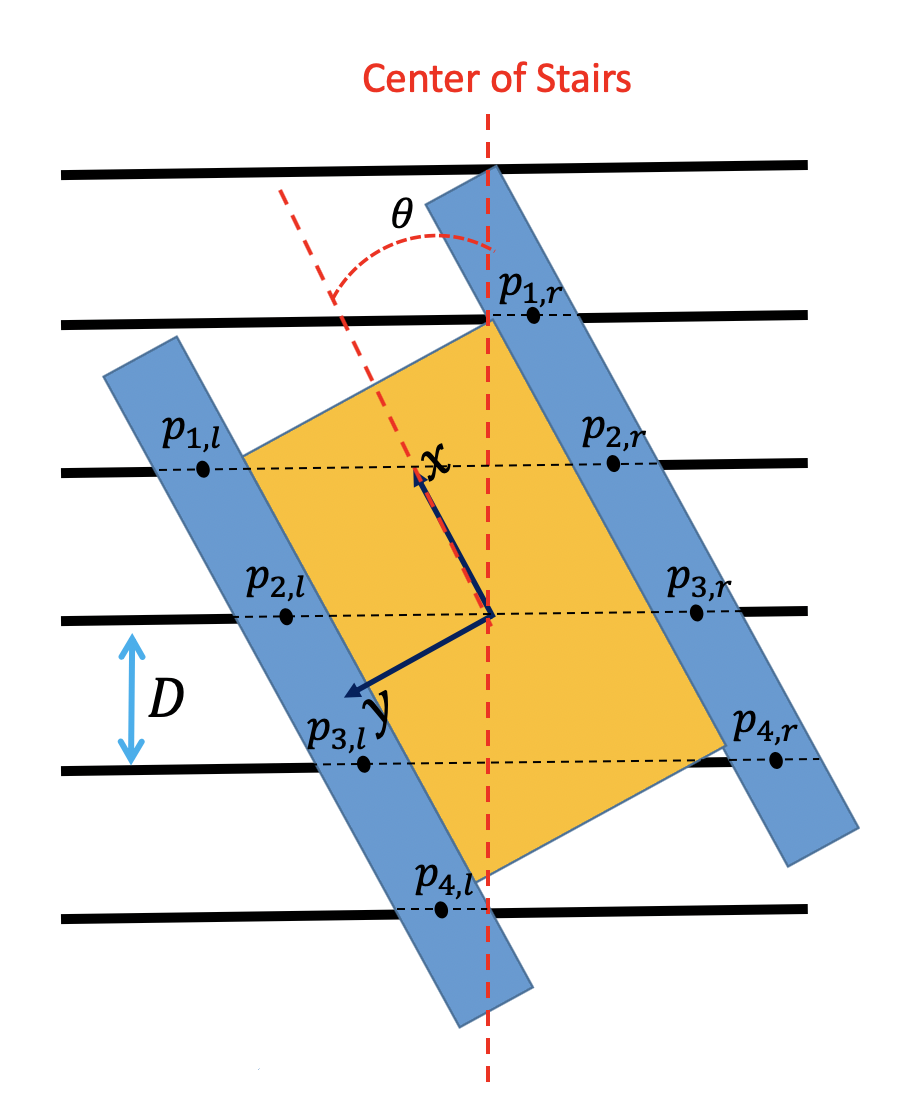}}
\caption{Geometry of a tracked robot on stairs}
\label{fig:stairs_fig}
\vskip -0.1 true in
\end{figure}

\noindent Next we consider the practically important case where the tracked vehicle climbs (or descends) a set of stairs. The tracks are assumed to contact just the lip of the stair (e,g., where the riser meets the tread). We assume that the stair edges are separated by a uniform distance $D$, and that the stairs are inclined at angle $\phi$ with respect to the horizontal.   We model the stair as a line contact, much like the grouser case.  Since the line of contact need not be aligned with body frame, the dissipation function depends on the body's orientation with respect to the stairs, see Fig. \ref{fig:stairs_fig}.

If the sum of the frictional reaction forces at the contacts are less in magnitude than the gravitational force acting on the body, the vehicle will accelerate in the direction of gravity, violating the quasi-static approximation. We assume that this is not the case.

In addition to the body orientation dependence, the power dissipation function also includes a slope-dependent term that captures the effects of gravity on vehicle motion: 
\begin{equation*}
\begin{aligned}
&\mathcal{D}_{stairs}(\dot{x}, \dot{y}, \dot{\theta}, \theta, S_r, S_l) =  \sum \limits_{n=1}^{N_c}\int \limits_{-T}^{T} \alpha_{n,r} \Bigg[(\dot{x} + (W - r_y\sec{\theta})\dot{\theta}- \\ & S_r)^2 + 
(\dot{y} + (p_{n,r} + r_y\tan{\theta})\dot{\theta})^2\Bigg]^\frac{1}{2} \mathrm{d}r_y
+ \, \sum \limits_{n=1}^{N_c} \int \limits_{-T}^{T}\alpha_{n,l}\Bigg[\Big(\dot{x} - \\ & (W+l_y\sec{\theta})\dot{\theta}-S_l\Big)^2 + (\dot{y} +(p_{n,l} + l_y\tan{\theta})\dot{\theta})^2\Bigg]^\frac{1}{2}\mathrm{d}l_y
\end{aligned}
\end{equation*}
Where $\alpha_{n,r} = \mu N(p_{n,r}) + mg\sin\phi$ and $\alpha_{n,l} = \mu N(p_{n,l}) + mg\sin\phi$ and $N(p_{n,r/l})$ is the normal force at $p_{n,r/l}$. As before, $p_{n,r/l}$ is the position ($x$-component) of the $n^{th}$ contact between the right/left track and the stairs.  
Once again, we calculate the normal forces by doing a moment balance. The normal forces change based on where the stairs contact the tracks. We assume the friction coefficients to be the same on all the stairs.

\section{Analysis} \label{sec:analysis}

The power dissipation method (\ref{eq:1}) predicts the tracked vehicle's body velocity for a given control input (track speeds). The vehicle motion on any terrain is obtained by minimizing $ \mathcal{D}(\dot{x}, \dot{y}, \dot{\theta}, \theta, S_r, S_l)$ with respect to the body velocities. It is generally not possible to find a closed form algebraic solution to this minimization problem. While the minimization described above is not burdensome, it is useful to express the state variables as a function of the control variables ($S_r, S_l$) for a fixed vehicle geometry. We approximate the dissipation function using least squares regression (see Appendix).


\begin{table}[h!]
\renewcommand{\arraystretch}{1.0}
\caption{Dimensions of the Rover Robotics Flipper}
\label{dimensions}
\centering
\begin{tabular}{c|c|c|c}
\hline
\bfseries Length-2L& \bfseries Width-2W & \bfseries Track Width-2T & \bfseries Grouser Pitch-D\\
\hline\hline
0.42 m& 0.27 m& 0.06 m& 0.04 m\\
\hline
\end{tabular}
\vskip -0.1 true in
\end{table}

\subsection{Flipper robot on flat ground}\label{ssec:analysis_flat}
In preparation for the experiments described in the next section, here we apply the principles described above to a model of the {\em Rover Robotics Flipper} tracked vehicle. The Flipper Rover has the dimensions given in Table \ref{dimensions}.

On flat ground, we use a $2^{nd}$-order approximation of the dissipation function. Using this approximation, a closed form approximation of the vehicle kinematics can be found by taking the partial derivative of the dissipation function w.r.t. $\dot{q}$ and setting it to $0$. This gives (with coefficients truncated to 3 decimal places), 
\begin{equation*}
{\small
    \begin{bmatrix} \dot{x} \\ \dot{y} \\ \dot{\theta} \end{bmatrix} 
     \approx  
    \begin{bmatrix}
    0.5 & 0.5  \\ 0 & 0\\ 1.5 & -1.5
    \end{bmatrix}
    \begin{bmatrix}
    S_r  \\ S_l
    \end{bmatrix}}
\end{equation*}

\subsection{Flipper robot on flat ground - with grousers}
For grousers that are about 0.04~m apart (the flipper robot dimensions) the least-squares fit yields the model: 
\begin{equation}{ \label{eq:flipper_kinematics}}
{\small
    \begin{bmatrix} \dot{x} \\ \dot{y} \\ \dot{\theta} \end{bmatrix} \approx
    \begin{bmatrix}
    0.5 & 0.5  \\ 0 & 0\\ 1.27  & -1.27
    \end{bmatrix}
    \begin{bmatrix}
    S_r  \\ S_l
    \end{bmatrix}}
\end{equation}

\subsection{Flipper robot on stairs}
\begin{figure}[htbp]
\centerline{\includegraphics[width=\columnwidth]{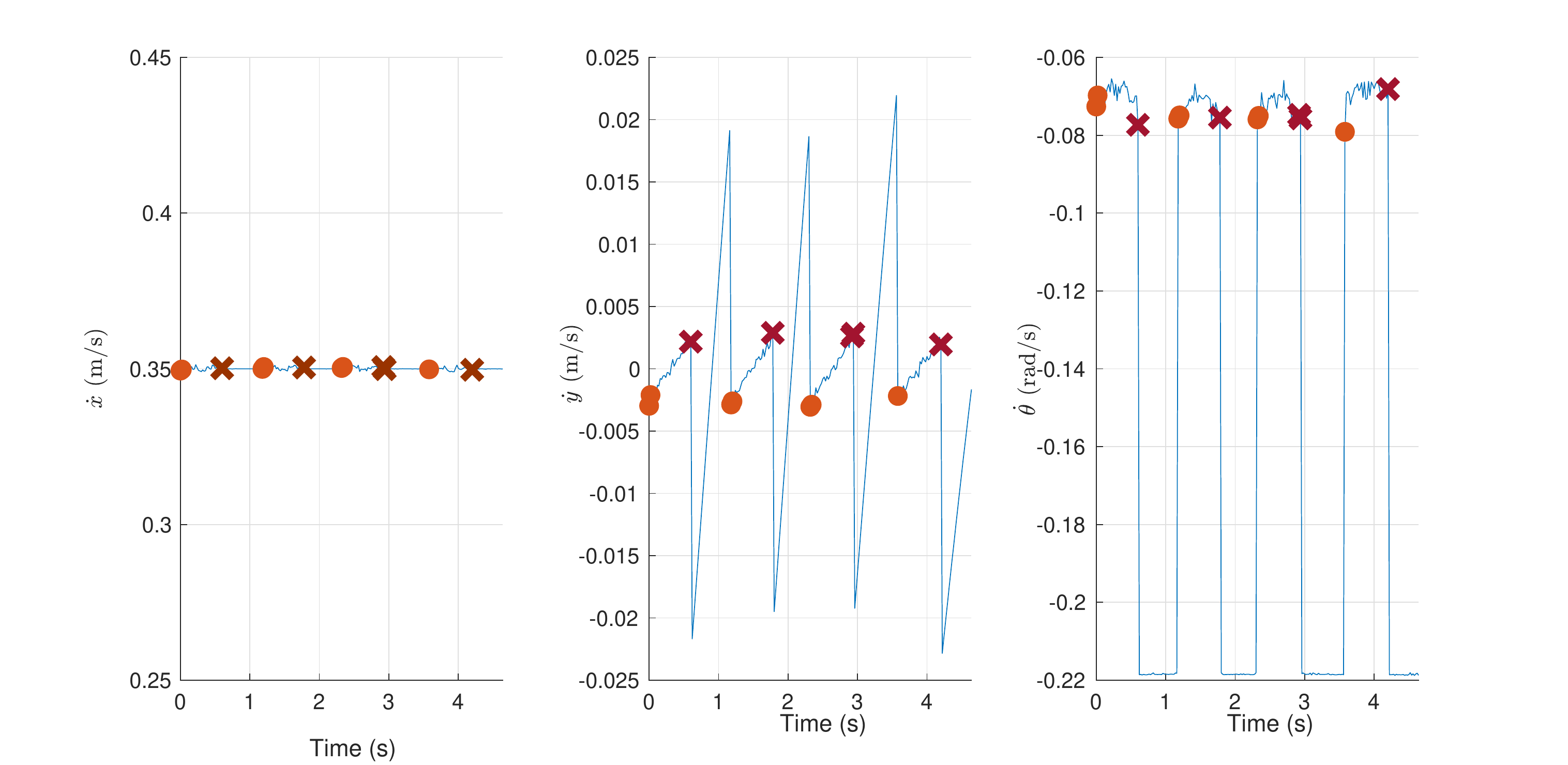}}
\caption{Simulated PDM body velocity prediction on stairs for fixed control inputs $(S_r, S_l) = (0.32, 0.38)m/s$. The {\em orange} o's denote points where a stair newly contacts a track, while {\em maroon} x's denote track loss of contact with a stair. }
\label{fig:stairs_switching}
\vskip -0.1 true in
\end{figure}

On stairs, our model predicts that vehicle motion depends upon the body orientation, $\theta$, with respect to the stairs and the locations of the tread contacts with the stairs. If we know the robot's initial configuration, we can predict its path on the stairs given $(S_r, S_l)$. 
It is important to note that the PDM predicts {\em discontinuous velocities} when tracks make and break contact with the stairs. The tracked vehicle's kinematics are a {\em hybrid systems model} that can be divided into two regions of continuous kinematics. One region occurs when the vehicle comes in contact with a new stair, till it loses contact with a stair (phase 1, in Fig. \ref{fig:stairs_switching}, o $\xrightarrow{}$ x). The other region is described by the track losing a stair contact till coming in contact with a new stair (phase 2, in Fig. \ref{fig:stairs_switching}, x $\xrightarrow{}$ o). To describe the switching surfaces between these regions, we introduce a new variable, $d$, the point of contact of the {\em first} stair with the track, along the vehicle x-axis. When the tracks contact a new stair, $d = 0$. Hence, this condition describes a transition from phase 2 to phase 1. It can be shown that the transition from phase 1 to phase 2 happens when $d = 2L \mod D\sec\theta$. 
\section{Experiments}  \label{sec:experiments}



This section presents experiments with the Flipper robot, see Table \ref{dimensions}, moving on flat ground and on stairs.  We compare the data against the PDM model derived earlier, and simple differential drive models used in the literature. In the flat ground experiments, we aim to follow a prescribed circular path and compare the PDM results with other simple models. On stairs, we first compare the path predicted by PDM for a fixed track input with that predicted by the other models. Finally, we take a closer look at the velocity on stairs to verify the switching behavior in the velocity predicted by the PDM model. 

\begin{figure}[htbp]
\centerline{\includegraphics[width=55mm,scale=0.5]{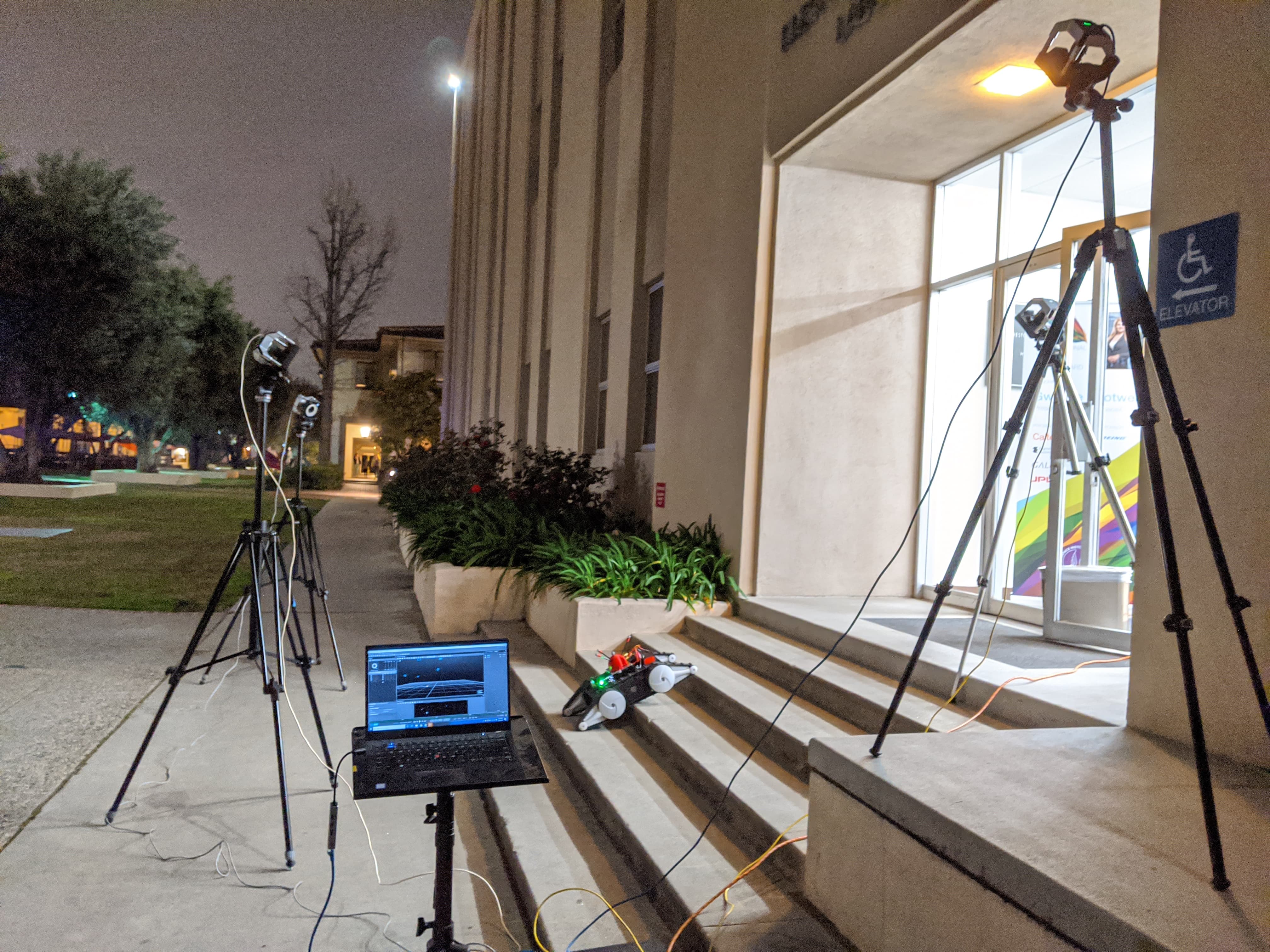}}
\caption{Photograph of the stairs and motion capture system used in the "stair" experiments described in this section.}
\label{fig:motioncapture_setup}
\vskip -0.2 true in
\end{figure}

In each experiment, the Flipper's rigid body motions and positions were recorded using an Optitrack motion capture system (120 Hz).  Fig. \ref{fig:motioncapture_setup} shows the motion capture system and the stairs used in our experiments.
The stair experiments additionally also used an onboard Intel Realsense T265 tracking camera (200 Hz) to record odometry. 

\subsection{Flat ground}
\begin{figure}[htbp]
\vskip -0.1 true in
\centering
\begin{subfigure}{0.5\textwidth}{
        \centering
        \includegraphics[width=80mm,scale=1.5]{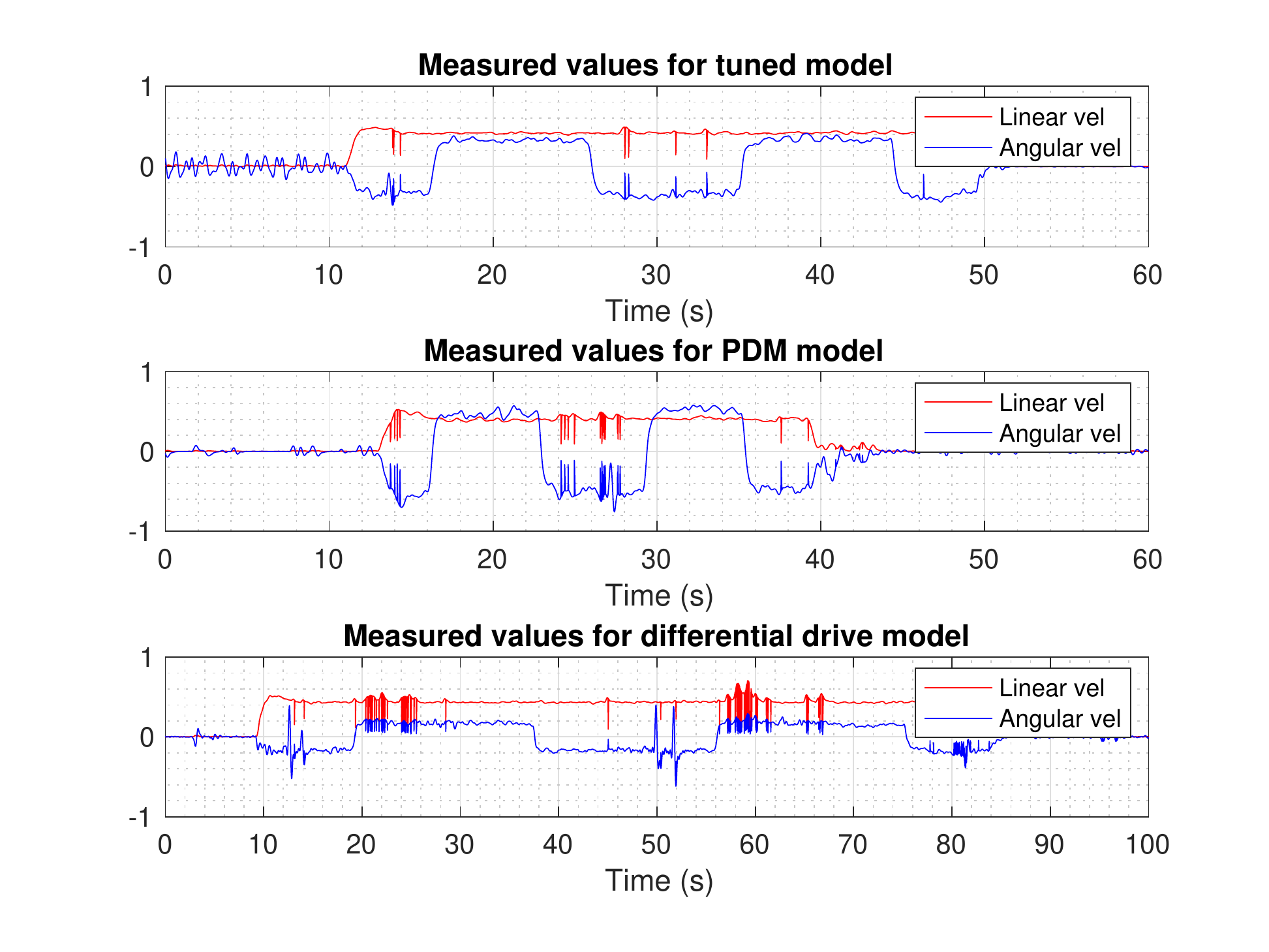}
        \vskip -0.1 true in
        \caption{Measured velocity}
        \label{fig:comparison_vel}}
\end{subfigure}\\
\begin{subfigure}{0.5\textwidth}{
        \centering
        \includegraphics[width=70mm,scale=1.0]{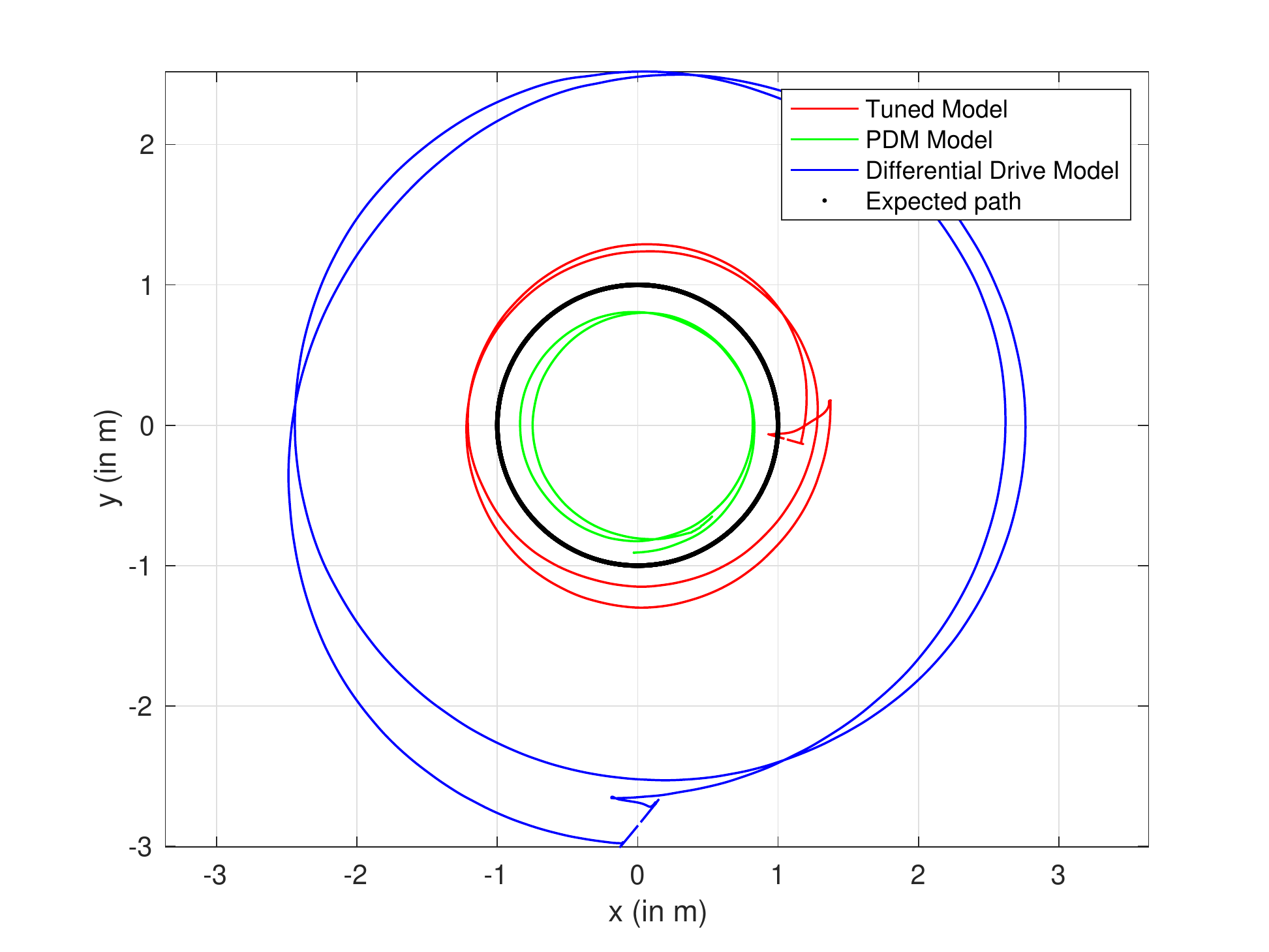}
        \vskip -0.1 true in
        \caption{Trajectory followed}
        \label{fig:comparison_traj}}
\end{subfigure}
\caption{Flat ground: Comparison of paths produced under different motion models when the commanded linear velocity $= 0.5$ m/s and angular velocity $= -0.5$ rad/sec}
\label{fig:comparison}
\vskip -0.1 true in
\end{figure}

\noindent 
We tested the Flipper robot's ability to track a 2 meter diameter circular path on flat, hard ground.  The Flipper was commanded via different motion models (1) a differential drive model: $\dot{x} = (S_r + S_l) / 2, \, \dot{\theta} = 2.38(S_r - S_l)$; (2) a "tuned" kinematic model provided by Rover Robotics  (using system identification): $\dot{x} = (S_r + S_l) / 2, \, \dot{\theta} = 1.37(S_r - S_l)$; and (3) the power dissipation model, with grousers at the same spacing as found on the Flipper's tracks: $\dot{x} = (S_r + S_l) / 2, \, \dot{\theta} = 1.27(S_r - S_l)$.  The tuned model includes traction coefficients and is specialized to this robot.  We calculated the vehicle forward ($0.5$ m/sec) and angular ($0.5$ rad/sec) velocities needed to follow the  circle, and then used the different models to calculate the track speeds needed to follow the circular path. The actual linear and angular velocity attained by the robot were derived from motion capture data.  With an open loop procedure, the errors between the commanded motion and actual motion are indicative of the errors in the underlying physics model. Fig. \ref{fig:comparison_vel} shows the velocities attained under the different models.  Fig. \ref{fig:comparison_traj} shows the actual robot paths (all paths are plotted with respect to a common circular center location) under the three different models. 


The measured forward velocity is observed to be a little lower than predicted in all three models. This is likely due to additional power losses due to slippage between the tracks and ground and due to track deformations. The measured angular velocity, however, is the closest to desired for the PDM model. The PDM model provides great accuracy (similar to the tuned model) despite it not using any experimental identification.

\subsection{Stairs}

\begin{figure}[htbp]
\vskip -0.1 true in
\centerline{\includegraphics[width=\columnwidth]{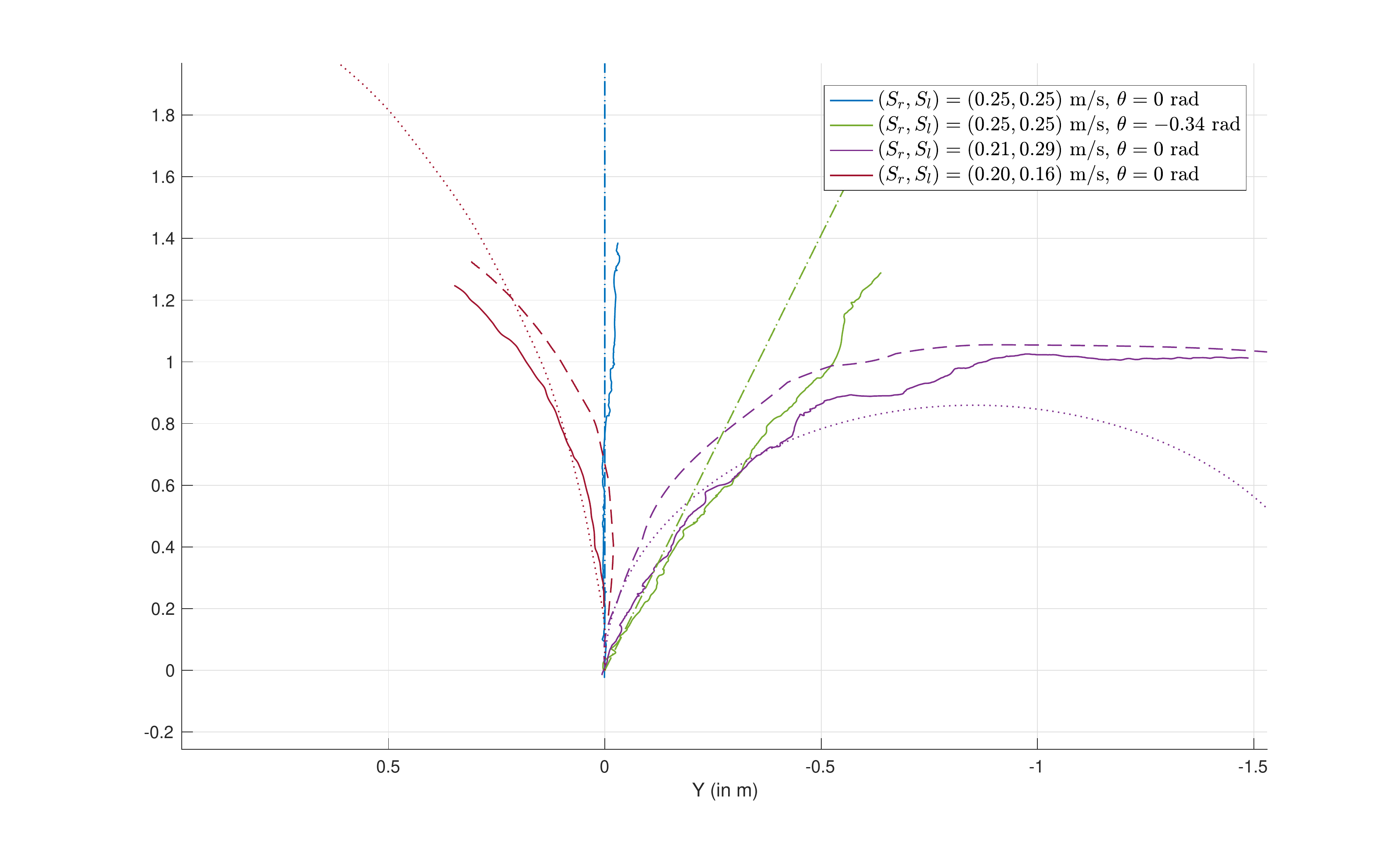}}
\vskip -0.1 true in
\caption{Trajectory on stairs: PDM prediction (dashed line), differential drive prediction (dotted line), experimental data (thick line).}
\label{fig:stair_experiment}
\end{figure}

We checked model validity by predicting how the vehicle should move on stairs for fixed inputs $(S_r, S_l)$, and at different starting orientations of the vehicle with respect the main stair axis. We  commanded the Flipper with these same track speeds and recorded the actual trajectory. Fig. \ref{fig:stair_experiment} shows four such trajectories (the plot legend provides the input speeds $(S_r, S_l)$ and initial orientation $\theta$). Note that the trajectories are symmetric, i.e., we get the same trajectories but in the opposite direction if we interchange  track speeds.  


The PDM model predicts that the vehicle will drive straight when the track speeds are equal, regardless of the robot's initial orientation. This behavior is seen in the experiments, see Fig. \ref{fig:stair_experiment}. With unequal track speeds, the robot will move with nonzero angular velocity.  In this case the stair model predicts the path followed with some accuracy. In the specific case shown in Fig. \ref{fig:stair_experiment} where $(S_r, S_l) = (0.21, 0.29)$ m/s, the model predicts that the robot will make a $90^o$ turn before reaching the end of the stairs, i.e., it will never completely climb the stairs.  This is exactly the behavior seen in the associated experiment.   The differential drive model cannot predict the behavior with such detail as it does not take into account the interactions between the tracks and stairs. It predicts a circular motion on the stair, which is not observed.

\subsection{Switching Behavior}

\begin{figure}[htbp]
\centerline{\includegraphics[width=75mm,scale=0.5]{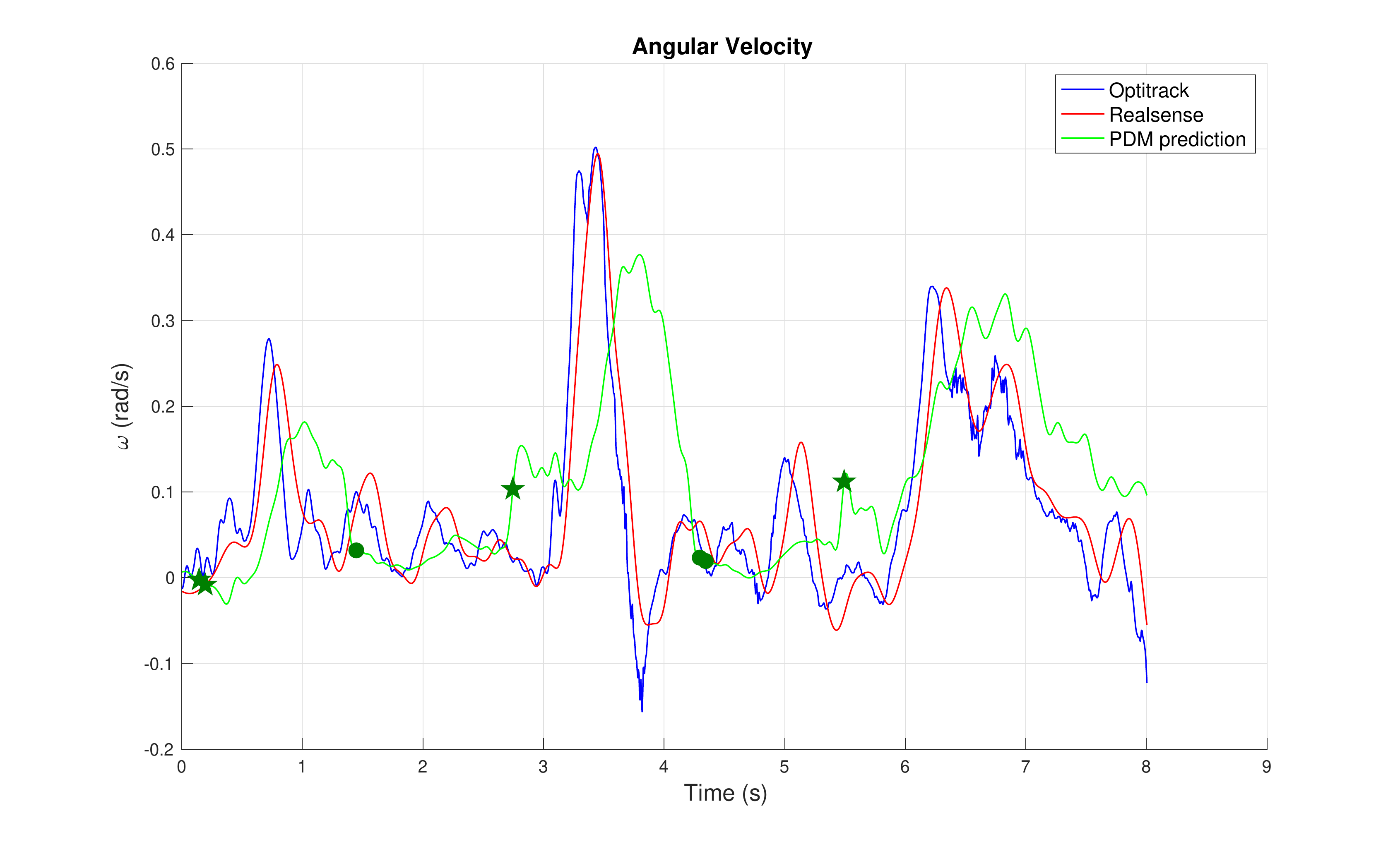}}
\caption{Switching behavior on stairs: Angular velocity for control inputs $(S_r, S_l) = (0.2, 0.16)m/s$. The o's denote points where a stair newly contacts a track, while $\star$'s denote loss of track contact with a stair.}
\label{fig:switching}
\vskip -0.1 true in
\end{figure}

In this subsection, we take a closer look at the angular velocity of the robot for one of the stair experiments from the previous subsection: $(S_r, S_l) = (0.2, 0.16)m/s$. A novel prediction of our theory is that nearly discontinuous switching-like behavior should be seen in Flipper angular velocities when the vehicle tracks make and break stair contact. To check these  predictions, we measured vehicle odometry on stairs using two methods: the Optitrack system described earlier and a Intel Realsense T265 tracking camera, which estimates visual odometry at 200 Hz.  For better prediction, the actual track speeds measured by the encoders on the robot were used as inputs to the PDM model. The robot's initial configuration is aligned with the axis of the stairs and the back of both tracks touch a stair lip. Fig. \ref{fig:switching} shows the switching behavior in the angular velocity as the robot moves over three stairs. Consistent with our analysis in the earlier section, the angular velocity increases when the tracks are in contact with fewer stairs and vice-versa. 



\section{Conclusions}

\noindent
This paper novelly applies the power dissipation modeling method to tracked robots.  This method leads to simple and efficient quasi-static models that can capture the surprisingly subtle effects of grousers on vehicle motion, as well as interaction with stairs.  Experiments show that this model provides better motion predictions than other kinematic models for driving on flat ground and for climbing stairs.

There are many avenues for future work.  Our models provide new insight into state switching behavior on stairs. Hence, we can pose the quasi-static mechanics of tracked vehicles as a switched system. 
An obvious next step is to develop feedback controllers based on these models in order to improve trajectory tracking performance, especially on stairs.  Intuitively, this result suggests that feedback controllers should estimate the vehicle's phase on the stairs and the cadence of the discontinuities, so that better stabilization to a specified stair-climbing trajectory can be realized.

This paper uses a simplistic Coulomb friction model.  But the power dissipation methodology allows for other friction loss models, such as grouser/soil interaction models \cite{bekker_off_1962,bekker_introduction_1969}. This extension will provide useful motion prediction on soil.  More importantly, our work supports a family of motion models on different terrain types that can be integrated within a hybrid control framework to allow for adaptive behavior across different terrain situations.

Finally, this paper assumes regular contact between the tracks and the ground or stairs.  On uneven terrain it is practically impossible to know the exact state of contact between the tracks and the ground.  The PDM framework should support an approach that has been taken in the quasi-static pushing literature \cite{huang_exact_2017}: bounds on the possible vehicle motions can be derived from the track geometry.  

\appendix 
Since the power dissipation function is always positive, we can use a {\em sum of squares} formulation \cite{parrilo2003semidefinite} to approximate the dissipation function as a polynomial of order $2k$. These polynomial coefficients can be found via least squares regression. The matrix, $A$, of these coefficients is positive semidefinite: $A \succeq 0$. Let $\hat{\mathcal{D}}$ denote the sum of squares polynomial:
\begin{equation}
    \hat{\mathcal{D}}(\dot{q}, u, A) = \nu^T A \nu
\end{equation}
where,
\begin{equation*}
    \dot{q} = \begin{bmatrix} \dot{x} & \dot{y} & \dot{\theta}
    \end{bmatrix}^T, \quad u = \begin{bmatrix} S_r & S_l
    \end{bmatrix}^T
\end{equation*}
\begin{equation*}
    \nu = \begin{bmatrix} 1 & \dot{q}^T &\dot{x}\dot{y}& \dot{x}\dot\theta& \dotsc& (\dot{q}^k)^T
    \end{bmatrix}^T
\end{equation*}
\begin{equation*}
    A \in \mathbb{R}^{m\times m}, \quad m = \binom{d + k}{k}.
\end{equation*}
Here, $d$ is the number of state and control variables, i.e, $d = 5$.
for the planar vehicle model. The polynomial fitting error is,
\begin{equation}
    E(\dot{q}, u, A) = \hat{\mathcal{D}}(\dot{q}, u, A) - \mathcal{D}_{min}(\dot{q}, u)
\end{equation}
where, $\mathcal{D}_{min}(\dot{q}, u)$ is the minimum value of $\mathcal{D}$ obtained from (\ref{eq:minD}).
The polynomial fitting error minimization is written as
\begin{equation} \label{eq:SOS}
\begin{aligned}
    \min_{A} \quad &\sum_{i = 1}^n E(\dot{q}_i, u_i, A)E^T(\dot{q}_i, u_i, A) \quad \mathrm{s.t.} \\
    & \frac{\partial}{\partial \dot{q}}\hat{\mathcal{D}}(\dot{q}_i, u_i, A) = 0 \quad  \forall \, i\in\{1,2,\dotsc,n\} \\
    & A \succeq 0\ .
\end{aligned}
\end{equation}
The above minimization finds a polynomial fit of the dissipation function. All of the data points are the minima of the dissipation function for a given control input (by virtue of being the PDM solutions). The gradient constraint ensures that we fit a dissipation function that has the same minima.

\section*{Acknowledgements}
The authors would like to thank Amanda Bouman, Matthew Anderson, and Mohamadreza Ahmadi for their insights and help in experimental setup and testing.

\bibliographystyle{IEEEtran}
\bibliography{IEEEexample.bib, TrackedVehicle.bib}

\begin{thebibliography}{10}
\providecommand{\url}[1]{#1}
\csname url@samestyle\endcsname
\providecommand{\newblock}{\relax}
\providecommand{\bibinfo}[2]{#2}
\providecommand{\BIBentrySTDinterwordspacing}{\spaceskip=0pt\relax}
\providecommand{\BIBentryALTinterwordstretchfactor}{4}
\providecommand{\BIBentryALTinterwordspacing}{\spaceskip=\fontdimen2\font plus
\BIBentryALTinterwordstretchfactor\fontdimen3\font minus
  \fontdimen4\font\relax}
\providecommand{\BIBforeignlanguage}[2]{{%
\expandafter\ifx\csname l@#1\endcsname\relax
\typeout{** WARNING: IEEEtran.bst: No hyphenation pattern has been}%
\typeout{** loaded for the language `#1'. Using the pattern for}%
\typeout{** the default language instead.}%
\else
\language=\csname l@#1\endcsname
\fi
#2}}
\providecommand{\BIBdecl}{\relax}
\BIBdecl

\bibitem{brunner_motion_2012}
M.~Brunner, B.~Bruggemann, and D.~Schulz, ``Motion planning for actively
  reconfigurable mobile robots in search and rescue scenarios,'' in
  \emph{{IEEE} {Int.} {Symp.} {Safety}, {Security}, and {Rescue} {Robotics}},
  College Station, TX, 2012, pp. 1--6.

\bibitem{bekker_off_1962}
M.~Bekker, \emph{Off-the-Road Locomotion}.\hskip 1em plus 0.5em minus
  0.4em\relax Univ. Michigan Press, 1960.

\bibitem{bekker_introduction_1969}
------, \emph{Introduction to Terrain-Vehicle Systems}.\hskip 1em plus 0.5em
  minus 0.4em\relax Univ. Michigan Press, 1969.

\bibitem{wong_general_2001}
J.~Y. Wong and C.~F. Chiang, ``A general theory for skid steering of tracked
  vehicles on firm ground,'' \emph{J. Automobile Engineering}, vol. 215, no.~3,
  pp. 343--355, Mar. 2001.

\bibitem{kitano_analysis_1977}
M.~Kitano and M.~Kuma, ``An analysis of horizontal plane motion of tracked
  vehicles,'' \emph{J. Terramechanics}, vol.~14, no.~4, pp. 211--225, 1977.

\bibitem{jozaki_steerability_1979}
H.~Jozaki and M.~Kitano, ``Steerability of {Tracked} {Vehicle} on {SOft}
  {Soil}: ({Part} 1): {Theoretical} {Analysis},'' \emph{J. Agriculatural
  Machinery Society, Japan}, vol.~40, no.~4, pp. 509--515, 1979.

\bibitem{baladi_analysis_1981}
G.~Baladi and B.~Rohani, ``\BIBforeignlanguage{en}{Analysis of {Steerability}
  of {Tracked} {Vehicles}: {Theoretical} {Predictions} versus {Field}
  {Measurements}},'' in \emph{\BIBforeignlanguage{en}{7th {Congress} {Int}.
  {Society} for {Terrain}-{Vehicle} {Syst.s}}}, Calgary, CA, 1981, p.~48.

\bibitem{wang_design_1990}
G.~Wang, S.~Wang, and C.~Chen, ``Design of turning control for a tracked
  vehicle,'' \emph{IEEE Control Syst.s Mag.}, vol.~10, no.~3, pp. 122--125, apr
  1990.

\bibitem{gonzalez_localization_2009}
R.~Gonzalez, F.~Rodriguez, J.~Guzman, and M.~Berenguel, ``Localization and
  control of tracked mobile robots under slip conditions,'' in \emph{{IEEE}
  {Int.} {Conf.} {Mechatronics}}.\hskip 1em plus 0.5em minus 0.4em\relax
  Malaga, Spain: IEEE, 2009, pp. 1--6.

\bibitem{jingang_yi_kinematic_2009}
{Jingang Yi}, {Hongpeng Wang}, {Junjie Zhang}, {Dezhen Song}, S.~Jayasuriya,
  and {Jingtai Liu}, ``Kinematic {Modeling} and {Analysis} of {Skid}-{Steered}
  {Mobile} {Robots} {With} {Applications} to {Low}-{Cost}
  {Inertial}-{Measurement}-{Unit}-{Based} {Motion} {Estimation},'' \emph{IEEE
  Trans. Robotics}, vol.~25, no.~5, pp. 1087--1097, 2009.

\bibitem{wei_yu_analysis_2010}
{Wei Yu}, O.~Chuy, E.~Collins, and P.~Hollis, ``Analysis and {Experimental}
  {Verification} for {Dynamic} {Modeling} of {A} {Skid}-{Steered} {Wheeled}
  {Vehicle},'' \emph{IEEE Trans. Robotics}, vol.~26, no.~2, pp. 340--353, Apr.
  2010.

\bibitem{janarthanan_lateral_2011}
B.~Janarthanan, C.~Padmanabhan, and C.~Sujatha,
  ``\BIBforeignlanguage{en}{Lateral dynamics of single unit skid-steered
  tracked vehicle},'' \emph{\BIBforeignlanguage{en}{Int. J. Automotive
  Technology}}, vol.~12, no.~6, pp. 865--875, Dec. 2011.

\bibitem{gonzalez_autonomous_2014}
R.~González, F.~Rodríguez, and J.~L. Guzmán, \emph{Autonomous {Tracked}
  {Robots} in {Planar} {Off}-{Road} {Conditions}}.\hskip 1em plus 0.5em minus
  0.4em\relax Springer Int. Pub., 2014, vol.~6.

\bibitem{pentzer_model-based_2014}
J.~Pentzer, S.~Brennan, and K.~Reichard, ``Model-based {Prediction} of
  {Skid}-steer {Robot} {Kinematics} {Using} {Online} {Estimation} of {Track}
  {Instantaneous} {Centers} of {Rotation}: {Model}-based {Prediction} of
  {Skid}-steer {Robot} {Kinematics},'' \emph{J. Field Robotics}, vol.~31,
  no.~3, pp. 455--476, 2014.

\bibitem{tianyou_guo_simplified_2013}
{Tianyou Guo} and {Huei Peng}, ``A simplified skid-steering model for torque
  and power analysis of tracked small unmanned ground vehicles,'' in
  \emph{{American} {Control} {Conf.}}, Washington, DC, Jun. 2013, pp.
  1106--1111.

\bibitem{ZviShiller}
Z.~{Shiller}, W.~{Serate}, and M.~{Hua}, ``Trajectory planning of tracked
  vehicles,'' in \emph{IEEE Int. Conf. Robotics and Automation}, May 1993, pp.
  796--801 vol.3.

\bibitem{le_estimation_nodate}
A.~T. Le, D.~C. Rye, and H.~F. Durrant-Whyte, ``Estimation of {Back}-soil
  {Interactions} for {Autonomous} {Tracked} {Vehicles},'' p.~6.

\bibitem{yi_adaptive_2007}
J.~Yi, D.~Song, J.~Zhang, and Z.~Goodwin, ``Adaptive {Trajectory} {Tracking}
  {Control} of {Skid}-{Steered} {Mobile} {Robots},'' in \emph{{IEEE} {Int.}
  {Conf.} {Robotics} and {Automation}}, Rome, 2007, pp. 2605--2610, iSSN:
  1050-4729.

\bibitem{moosavian_experimental_2008}
S.~Moosavian and A.~Kalantari, ``Experimental slip estimation for exact
  kinematics modeling and control of a {Tracked} {Mobile} {Robot},'' in
  \emph{{IEEE}/{RSJ} {Int.} {Conf.} {Intel.} {Robots} and {Syst.s}}, Nice,
  2008, pp. 95--100.

\bibitem{dar_slip_2010}
T.~M. Dar and R.~G. Longoria, ``Slip estimation for small-scale robotic tracked
  vehicles,'' in \emph{{American} {Control} {Conf}}, Baltimore, MD, Jun. 2010,
  pp. 6816--6821.

\bibitem{martinez_approximating_2005}
J.~L. Martínez, A.~Mandow, J.~Morales, S.~Pedraza, and A.~García-Cerezo,
  ``Approximating {Kinematics} for {Tracked} {Mobile} {Robots},'' \emph{Int. J.
  Robotics Research}, vol.~24, no.~10, pp. 867--878, Oct. 2005.

\bibitem{morales_power_modeling_2009}
J.~{Morales}, J.~L. {Martinez}, A.~{Mandow}, A.~J. {Garcia-Cerezo}, and
  S.~{Pedraza}, ``Power consumption modeling of skid-steer tracked mobile
  robots on rigid terrain,'' \emph{IEEE Transactions on Robotics}, vol.~25,
  no.~5, pp. 1098--1108, 2009.

\bibitem{yugang_liu_track--stair_2009}
{Yugang Liu} and {Guangjun Liu}, ``Track--{Stair} {Interaction} {Analysis} and
  {Online} {Tipover} {Prediction} for a {Self}-{Reconfigurable} {Tracked}
  {Mobile} {Robot} {Climbing} {Stairs},'' \emph{IEEE/ASME Trans. Mechatronics},
  vol.~14, no.~5, pp. 528--538, Oct. 2009.

\bibitem{yalin_xiong_vision-guided_2000}
{Yalin Xiong} and L.~Matthies, ``Vision-guided autonomous stair climbing,'' in
  \emph{{IEEE} {Int.} {Conf.} {Robotics} and {Automation}}, San Francisco, CA,
  USA, 2000, pp. 1842--1847.

\bibitem{mourikis_autonomous_2007}
A.~I. Mourikis, N.~Trawny, S.~I. Roumeliotis, D.~M. Helmick, and L.~Matthies,
  ``Autonomous {Stair} {Climbing} for {Tracked} {Vehicles},'' \emph{Int. J.
  Robotics Research}, vol.~26, no.~7, pp. 737--758, Jul. 2007.

\bibitem{Steplight}
S.~{Steplight}, G.~{Egnal}, S.~{Jung}, D.~{Walker}, C.~{Taylor}, and
  J.~{Ostrowski}, ``A mode-based sensor fusion approach to robotic
  stair-climbing,'' in \emph{IEEE/RSJ Int. Conf. Intel. Robots and Syst.s},
  2000, pp. 1113--1118.

\bibitem{li_research_2014}
L.~Li, W.~Wang, D.~Wu, and Z.~Du, ``Research on obstacle negotiation capability
  of tracked robot based on terramechanics,'' in \emph{{IEEE}/{ASME} {Int.}
  {Conf.} {Advanced} {Intel.} {Mechatronics}}, Besacon, 2014, pp. 1061--1066.

\bibitem{minPower}
M.~A. {Peshkin} and A.~C. {Sanderson}, ``Minimization of energy in quasistatic
  manipulation,'' in \emph{IEEE Int. Conf. Robotics and Automation}, April
  1988, pp. 421--426.

\bibitem{WMR}
J.~Alexander and J.~Maddocks, ``On the kinematics of wheeled mobile robots,''
  \emph{Int. J. Robotic Research}, vol.~8, pp. 15--27, 10 1989.

\bibitem{PDM}
T.~Murphey and J.~Burdick, ``The power dissipation method and kinematic
  reducibility of multiple-model robotic systems,'' \emph{IEEE Trans.
  Robotics}, vol.~22, pp. 694 -- 710, 09 2006.

\bibitem{huang_exact_2017}
E.~Huang, A.~Bhatia, B.~Boots, and M.~Mason, ``Exact {Bounds} on the {Contact}
  {Driven} {Motion} of a {Sliding} {Object}, {With} {Applications} to {Robotic}
  {Pulling},'' in \emph{Robotics: {Science} and {Systems} {XIII}}, Jul. 2017.

\bibitem{parrilo2003semidefinite}
P.~A. Parrilo, ``Semidefinite programming relaxations for semialgebraic
  problems,'' \emph{Math. programming}, vol.~96, no.~2, pp. 293--320, 2003.

\end{thebibliography}

\vspace{12pt}
\color{red}

\end{document}